\documentclass[11pt,a4paper]{article}
\usepackage[margin=2.4cm]{geometry}
\usepackage{amsmath,amssymb,amsthm}
\usepackage{algorithm}
\usepackage{algpseudocode}
\usepackage{graphicx}
\usepackage{booktabs}
\usepackage{hyperref}
\usepackage{microtype}
\usepackage{parskip}
\usepackage{xcolor}
\usepackage{multirow}
\usepackage{array}

\hypersetup{colorlinks=true,linkcolor=blue,citecolor=blue,urlcolor=blue}

\newtheorem{definition}{Definition}

\title{\textbf{Optimizing Earth Observation Satellite Schedules\\
under Unknown Operational Constraints:\\
An Active Constraint Acquisition Approach}}

\author{
  Mohamed-Bachir Belaid\\[4pt]
  \small NILU, Climate and environmental research institute, Kjeller, Norway\\
  \small \texttt{bbel@nilu.no}
}
\date{}

\begin{document}
\maketitle

\begin{abstract}
Earth Observation (EO) satellite scheduling (deciding which imaging tasks
to perform and when) is a well-studied combinatorial optimization problem.
Existing methods typically assume that the operational constraint model is
fully specified in advance. In practice, however, constraints governing
separation between observations, power budgets, and thermal limits are
often embedded in engineering artefacts or high-fidelity simulators rather
than in explicit mathematical models.
We study EO scheduling under
\emph{unknown constraints}: the objective is known, but feasibility must
be learned interactively from a binary oracle. Working with a simplified
model restricted to pairwise separation and global capacity constraints,
we introduce Conservative Constraint Acquisition~(CCA), a domain-specific procedure
designed to identify justified constraints efficiently in practice while
limiting unnecessary tightening of the learned model. Embedded in the
\textsc{Learn\&Optimize} framework, CCA supports an interactive search
process that alternates optimization under a learned constraint model with
targeted oracle queries. On synthetic instances with up to 50~tasks and
dense constraint networks, L\&O improves over a no-knowledge
greedy baseline and uses far fewer main oracle queries than a two-phase
acquire-then-solve baseline (FAO). For $n\leq 30$, the average gap drops from
65--68\% (Priority Greedy) to 17.7--35.8\% using L\&O. At $n{=}50$, where the CP-SAT
reference is the best feasible solution found in 120~s, L\&O improves on
FAO on average (17.9\% vs.\ 20.3\%) while using 21.3 main queries instead
of 100 and about $5\times$ less execution time.
\end{abstract}

\textbf{Keywords:} Earth observation scheduling, constraint acquisition,
interactive optimisation, unknown constraints.

\vspace{1em}
\section{Introduction}
\label{sec:intro}

\subsection{The Problem}
\label{sec:plain}

An Earth Observation satellite orbits the planet and has a limited number of
opportunities to photograph ground targets during each pass.  Each target
has a \emph{time window} during which it is visible and a \emph{priority}
reflecting its importance.  The scheduler must decide which targets to
image and at which time slot, so as to maximize total priority while
respecting operational constraints.
Following the EO literature~\cite{wang2021twenty,lemaitre2002}, we refer
to this as a \emph{scheduling} problem.  


\paragraph{Running example.}
Consider a satellite passing over three ground targets.  We discretize the
orbit into time slots of, say, 30~seconds each.  Each target is visible
only during a short arc of the orbit:
\begin{center}
\small
\begin{tabular}{lccc}
\toprule
Target & Priority & Visibility window & Meaning \\
\midrule
$A$ & 10 & $W_A = \{5\}$       & visible only at slot 5 \\
$B$ & 8  & $W_B = \{6\}$       & visible only at slot 6 \\
$C$ & 4  & $W_C = \{18,\ldots,22\}$ & visible during slots 18--22 \\
\bottomrule
\end{tabular}
\end{center}
The \emph{visibility window} $W_j$ lists the slots during which
target~$j$ is geometrically visible from the satellite.  Windows are
short (typically a few slots) because the satellite moves at ${\sim}7$~km/s
and any ground point is in view for at most a few minutes.

To photograph a target, the satellite must point its camera at the correct
angle.  After imaging one target, it must rotate to a new
angle for the next, then waits for settlement.  This time imposes a
\textbf{separation constraint}: if imaging $A$ requires a large rotation
to reach $B$, the satellite needs several slots to complete the manoeuvre.
For instance, $\mathrm{sep}(A,B,3)$ means the angular distance between
$A$ and $B$ requires at least $3 \times 30 = 90$~seconds of rotation and
settling.  The required gap $\delta$ is pair-specific: it depends on the
angular distance between the two targets as seen from the spacecraft.

In the example, $A$ is at slot~5 and $B$ at slot~6, giving a gap of
$|5-6| = 1 < 3$: the satellite is still rotating when $B$'s window
arrives, so it cannot image both.  The best feasible schedule is
$\{A, C\}$ with total priority of 14.

\subsection{Why Constraints Could be Unknown}
\label{sec:oracle_motivation}

Every existing EO scheduler takes
the constraint model as given.  In practice, this assumption breaks for
three reasons.

\textbf{Constraints live in engineering artefacts, not in models.}
The minimum separation time between observations depends on the
satellite's attitude control system (reaction wheels, star trackers,
firmware) and is typically specified in engineering margin documents, not
as a mathematical formula.  The power budget per orbit depends on battery
state-of-health, thermal conditions, and duty-cycle history.

\textbf{Constraints change over the mission lifetime.}
A firmware update may alter stabilization times. For example, battery degradation
tightens power budgets, a reaction wheel failure changes the rotation
capabilities,...  Each change invalidates the previous model and requires
manual re-derivation.

\textbf{For new missions, no validated model exists.}
The engineering team often has a high-fidelity simulator that can evaluate
any proposed plan, but extracting an explicit constraint model from the
simulator is expensive and error-prone.

These observations motivate the use of a \textbf{feasibility oracle}: a
system that evaluates any proposed schedule and returns \textit{yes} or
\textit{no}, without revealing which constraint was violated.  The natural
instantiation is a \textbf{high-fidelity spacecraft simulator}~\cite{aeosformer2025},
which models attitude dynamics, power subsystems, thermal behaviour,
and all onboard constraints.  Given a
proposed plan, the simulator propagates the full spacecraft state and
reports whether all hardware limits are satisfied. The oracle could also be an \textbf{operational
validation tool}, i.e.\ software that checks candidate schedules against flight
rules.  In both cases the oracle embodies the full
constraint model implicitly but does not expose it.

\paragraph{Why the oracle gives only yes/no.}
A natural question is why the oracle cannot simply report \emph{which}
constraint was violated, which would make acquisition trivial.  The reason
is that the oracle does not represent constraints in the planner's
language.  A spacecraft simulator propagates continuous physical
state (e.g., attitude quaternions, reaction wheel momentum, battery charge,
thermal loads,...) through coupled differential equations.  When a schedule
fails, the simulator reports a \emph{physical symptom}: ``reaction wheel
momentum exceeded safe margin at $t{=}47.3$\,s'' or ``battery
state-of-charge dropped below 20\% during slots 12--18.''  Translating
this into a discrete constraint like $\mathrm{sep}(A,B,3)$
requires understanding the relationship between rotation angles, reaction
wheel dynamics, and the specific task pair that caused the problem.
Moreover, a single
rejection may reflect multiple simultaneous violations (e.g.\ a thermal
overload caused by three observations jointly, not by any pair), and
operational validation tools are often black-box systems certified for
correctness, not for diagnostic output.

Even when simulators produce partial diagnostics (e.g.\ ``power
budget exceeded in window [10,\,25]''), the symptom does not specify the
constraint parameters: is it $\mathrm{cap}(2,15)$ or $\mathrm{cap}(3,20)$?
The mapping from physical symptoms to constraint language remains
ambiguous.

Our approach learns constraints from these binary answers while
simultaneously optimizing the schedule.

\subsection{Scope and Contributions}
\label{sec:contributions}

\paragraph{A simplified constraint model.}
Real satellite operations involve dozens of constraint types: orbital
dynamics, cloud coverage, downlink scheduling, stereo pair coordination,
thermal limits, and more. We study a \textbf{simplified model} restricted
to two operationally dominant families: \emph{pairwise separation}
(minimum time gap between observations) and \emph{global capacity}
(maximum observations in a sliding time window). This restriction lets us
design a domain-specific acquisition procedure and study the interaction
between oracle-guided learning and optimization in a controlled setting.

\paragraph{Contributions:}
\begin{enumerate}
  \item We formulate EO Scheduling under Unknown Constraints (EOSP-UC), a
        simplified model in which the feasibility constraints are hidden
        behind a binary oracle (Section~\ref{sec:problem}).

  \item We introduce \textbf{Conservative Constraint Acquisition (CCA)}, a
        domain-specific acquisition procedure tailored to the
        separation\,/\,capacity structure of EOSP-UC. CCA is \emph{not} a
        generic constraint acquisition algorithm: it exploits the ordering
        structure within these two constraint families and does not
        generalize to arbitrary languages (Section~\ref{sec:method}).

  \item We embed CCA in the \textsc{Learn\&Optimize} framework~\cite{belaid2021},
        interleaving constraint acquisition with CP-SAT optimization so
        that the algorithm improves the schedule continuously while learning,
        and terminates as soon as an oracle-accepted proposal is found
        rather than waiting for full acquisition (Section~\ref{sec:method}).

  \item We evaluate on synthetic instances with $n \in \{10,20,30,40,50\}$
        and dense constraint networks. For $n\leq 30$, L\&O reduces the
        average gap from 65--68\% (Priority Greedy) to 17.7--35.8\%.
        At $n{=}50$, L\&O improves over FAO on average (17.9\% vs.\ 20.3\%)
        while using 21.3 main queries instead of 100 and about $5\times$
        less wall-clock time (Section~\ref{sec:experiments}).
\end{enumerate}


In summary, this is (to the best of our knowledge) the first application
of active constraint acquisition to EO scheduling and the first study
of EO optimisztion under unknown constraints.
\section{Background and Related Work}
\label{sec:related}

\subsection{EO Satellite Scheduling}
\label{sec:related_eo}

The EO scheduling problem was formally posed by Bensana et
al.~\cite{bensana1999} and defined comprehensively by Lema\^{i}tre et
al.~\cite{lemaitre2002} for the French Pl\'eiades programme.  The goal is
to maximise total observation profit subject to visibility windows,
transition constraints (minimum rotation and stabilization time between
observations), and onboard resource limits (memory, energy).  Agile
satellites can adjust their attitude along three axes, extending
observation windows but introducing sequence-dependent transition times
that make the problem NP-hard~\cite{wang2021twenty}.

Solution approaches span exact methods
(branch-and-bound~\cite{gabrel1997,chu2017bb}, MILP~\cite{bianchessi2007},
constraint programming~\cite{verfaillie2001cp,miralles2026cp}),
constructive and neighbourhood-search
heuristics~\cite{lemaitre2002,vasquez2001,xu2016priority,he2018improved,han2022sa,wang2019complex},
and deep reinforcement
learning~\cite{chen2019drl,zhao2020two,wei2021drl,herrmann2023rl}.
Wang et al.~\cite{wang2021twenty} survey 62~papers over 20~years and Ferrari
et al.~\cite{ferrari2024survey} extend the coverage to~2024.  Every
method assumes a fully specified constraint model.

\subsection{Optimization under Uncertain or Unknown Constraints}
\label{sec:related_uncertain}

A related but distinct line of work addresses optimization when
constraints are \emph{uncertain} rather than unknown.  Dong et
al.~\cite{dong2021cuc} study cloud job scheduling where capacity
constraints are stochastic and must be predicted from historical data.
Their predict-then-optimise framework (CUC) handles distributional
uncertainty but assumes the constraint \emph{structure} is known and only
the parameters are noisy.  In contrast, our setting assumes prior knowledge of the constraint language but not of the hidden active model. That is, the learner knows the relevant families of constraints and a candidate basis, but must discover which constraints are actually justified, and at what parameter values, through binary feasibility queries.

\subsection{Constraint Acquisition}
\label{sec:related_ca}

Constraint acquisition learns a constraint satisfaction or optimisation
model from data.  \textsc{Conacq}~\cite{conacq} performs passive learning
from labelled examples.  \textsc{QuAcq}~\cite{quacq} introduced active
acquisition via partial-assignment queries: given a negative example, its
FindScope and FindC subroutines identify one exact constraint from the
hidden model~$T$, maintaining the strong invariant $L \subseteq T$.
Extensions include multi-constraint acquisition~\cite{arcangioli2016},
time-bounded query generation~\cite{addi2018}, structure
exploitation~\cite{tsouros2019} and qualitative constraint acquisition~\cite{belaid2024}.  

In its original formulation, \textsc{Learn\&Optimize}~\cite{belaid2021}
joins constraint acquisition with optimization, maintaining upper and lower
bounds on the true optimum and allowing early termination when those bounds
converge.
Bessiere et al.~\cite{bessiere2014solve} also study solving without
an explicit model but via passive acquisition rather than interactive
optimization.  No prior work has applied constraint acquisition to EO
scheduling.
\section{Problem Formulation}
\label{sec:problem}

\begin{definition}[EOSP-UC: EO Scheduling under Unknown Constraints]
\label{def:eospuc}
Let $J = \{1,\ldots,n\}$ be a set of imaging tasks, each with priority
weight $w_j > 0$ and a \emph{visibility window} $W_j \subseteq
\{1,\ldots,H\}$, where $H$ is the planning horizon in discrete slots.  Decision variable
$x_j \in \{0\} \cup W_j$ assigns task~$j$ to a specific slot within its
window, or leaves it unscheduled ($x_j = 0$).  In constraint acquisition
terminology, the domain of each variable is $D(x_j) = \{0\} \cup W_j$.  For example, if
$W_A = \{3,4,5\}$, then $D(x_A) = \{0,3,4,5\}$: the scheduler can image
$A$ at slot~3, 4, or~5, or skip it.  The choice of slot matters because
it affects separation from other tasks.  The objective is:
\begin{equation}
  \max_x \sum_{j=1}^{n} w_j \cdot \mathbf{1}[x_j > 0].
  \label{eq:obj}
\end{equation}
A hidden constraint set~$T$ restricts which assignments are feasible.  An
oracle~$\mathcal{O}$ accepts any complete assignment~$e$ and returns
\textit{yes} if $e$ satisfies every constraint in~$T$, \textit{no}
otherwise.  The oracle does not indicate which constraint was violated.
This \emph{membership oracle} is the standard model in constraint
acquisition~\cite{bessiere2017ca,quacq,belaid2021}: it reflects the
practical reality that feasibility checkers (simulators, validation tools)
evaluate schedules holistically without decomposing failures into
individual named constraints (see Section~\ref{sec:oracle_motivation} for
a detailed justification).  
\end{definition}

\paragraph{Simplified constraint model.}
We restrict $T$ to two families:
\begin{itemize}
  \item \textbf{Separation} $\mathrm{sep}(i,j,\delta)$: if both tasks
        $i$ and $j$ are scheduled, their slots must satisfy
        $|x_i - x_j| \geq \delta$.  Physically, $\delta$ reflects the
        time the satellite needs to rotate between the pointing angles of
        targets $i$ and $j$ and re-stabilise (see
        Section~\ref{sec:plain});
  \item \textbf{Capacity} $\mathrm{cap}(k,w)$: at most $k$~tasks may be
        scheduled in any window of $w$~consecutive slots.  Physically,
        this models the power budget (the battery cannot sustain more
        than $k$~high-power imaging activations within a $w$-slot
        interval) or downlink bandwidth limits.
\end{itemize}
Window constraints are embedded in the variable domains.  This is a
deliberate simplification: real operations involve additional families
(see Section~\ref{sec:contributions}).

\paragraph{Constraint language.}
The language $\mathcal{L}$ consists of two constraint families:
$\mathrm{sep}(i,j,\delta)$ (pairwise separation) and $\mathrm{cap}(k,w)$
(global capacity).  The candidate basis $B$ is obtained by instantiating
$\mathcal{L}$: $\mathrm{sep}(i,j,\delta)$ for all task pairs with a range
of plausible $\delta$ values, plus $\mathrm{cap}(k,w)$ for a neighbourhood
of plausible $(k,w)$ values. We assume $T \subseteq B$: the candidate basis is rich enough to
contain every true constraint.

\section{Methodology}
\label{sec:method}

\subsection{Learn\&Optimize Framework}

We apply \textsc{Learn\&Optimize}~\cite{belaid2021}
(Algorithm~\ref{alg:lo}) to EOSP-UC. We first define notations.

\smallskip
\begin{tabular}{@{}ll@{}}
$H$ & planning horizon (number of discrete slots) \\
$T$ & hidden true model (unknown) \\
$\mathcal{L}$ & constraint language (families / templates) \\
$B$ & instantiated candidate basis derived from $\mathcal{L}$ \\
$L$ & learned constraint set maintained by CCA \\
$B_{ij}$ & separation candidates in $B$ for pair $(i,j)$ \\
$f_T^*$ & optimal value under the hidden true model $T$ \\
$f_L^*$ & optimal value under the learned model $L$ \\
$f_{L\cup B}^*$ & optimal value under the tightened model $L \cup B$ \\
$\mathcal{O}$ & binary oracle
\end{tabular}
\smallskip

The algorithm maintains two sets. The learned set $L$ starts empty and grows
as the oracle rejects proposals. The candidate basis $B$ starts as the
instantiated constraints (generated from $\mathcal{L}$) that are consistent
with a dummy oracle-feasible schedule, and then shrinks as YES answers rule
out incompatible candidates. Thus, $\mathcal{L}$ denotes the \emph{language},
$B$ the current \emph{candidate basis}, and $L$ the current \emph{learned
constraint model}.

In the ideal case, if CCA preserves a sound learning invariant, then
$\textsc{OptSol}(L,f)$ can be interpreted as an optimistic model-side value
and $\textsc{OptSol}(L \cup B,f)$ as a tightened model-side value. In the
current CCA-based implementation, however, the learned model may be
over-constrained (due to the conservative approach), so we do \emph{not} assume a fixed ordering between
$f_L^*$ and the true hidden optimum $f_T^*$ in general. Accordingly,
$f_L^*$ and $f_{L\cup B}^*$ should be interpreted as optima of constraint
models used by the algorithm, not automatically as certified bounds on
$f_T^*$.

\paragraph{Query generation.}
Each iteration solves $\textsc{OptSol}(L, f)$, that is, it maximizes $f$
subject only to the currently learned constraints. Because $L$ is only a
part for the hidden model $T$, the returned schedule may still violate
constraints enforced by the oracle.

\paragraph{Model-side optima.}
For analysis we distinguish three quantities:
\begin{align}
  f_T^*      &= \max \{ f(e) : e \models T \}, \\
  f_L^*      &= \max \{ f(e) : e \models L \}, \\
  f_{L\cup B}^* &= \max \{ f(e) : e \models L \cup B \}.
\end{align}
Here $f_T^*$ is the true hidden optimum, while $f_L^*$ and $f_{L\cup B}^*$
are optima of the learned and tightened constraint models, respectively.
Only under an additional soundness invariant should these be interpreted as
upper or lower bounds on $f_T^*$.

\begin{algorithm}[t]
\caption{\textsc{Learn\&Optimize} for EOSP-UC}
\label{alg:lo}
\begin{algorithmic}[1]
\Require Tasks $J$, objective $f$, language $\mathcal{L}$,
         oracle $\mathcal{O}$, cutoff $Q$
\Ensure Best oracle-feasible schedule found
\State $e_p \gets$ single highest-priority task (oracle-confirmed)
\State $B \gets \{c \in \mathrm{Inst}(\mathcal{L}) : c(e_p)=\text{true}\}$;\;
       $L \gets \emptyset$;\; $e^* \gets e_p$
\For{$q = 1, \ldots, Q$}
  \State $e \gets \textsc{OptSol}(L, f)$
  \State $v_L \gets f(e)$
  \If{$\mathcal{O}(e) = \textit{yes}$}
    \State $e^* \gets e$;\; $B \gets \{c \in B : c(e)=\text{true}\}$
    \State \textbf{break}
    \Comment{\textbf{Accepted-Opt($L$)}: current $\textsc{OptSol}(L,f)$ proposal is oracle-feasible}
  \Else
    \State $\textsc{CCA}(L, B, e, \mathcal{O})$
  \EndIf
  \If{$B = \emptyset$}
    \State \textbf{break}
    \Comment{\textbf{Basis exhausted}: no candidate constraints remain}
  \EndIf
  \State $e_{L\cup B} \gets \textsc{OptSol}(L \cup B, f)$
  \State $v_{L\cup B} \gets f(e_{L\cup B})$
  \If{$v_L = v_{L\cup B}$}
    \State \textbf{break}
    \Comment{\textbf{Model-side convergence}: $f_L^* = f_{L\cup B}^*$}
  \EndIf
\EndFor
\State $e_f \gets \textsc{OptSol}(L, f)$; if $f(e_f) > f(e^*)$ and $T$-feasible, set $e^* \gets e_f$
\State \Return $e^*$
\end{algorithmic}
\end{algorithm}
\subsection{Conservative Constraint Acquisition (CCA)}
\label{sec:ca}

CCA is a \textbf{domain-specific} procedure designed for the
separation\,/\,capacity structure of EOSP-UC.  It is \textbf{not} a
general-purpose constraint acquisition algorithm: unlike
\textsc{QuAcq}~\cite{quacq}, it does not implement FindScope or FindC,
and it does not generalise to arbitrary constraint languages.  For other
constraint families, \textsc{QuAcq} or its successors should be used.

The key idea is \emph{conservatism}.  When the oracle rejects a schedule,
CCA refines the current constraint model using domain-specific partial queries and
prunes dominated candidates from the basis. In an idealized setting this
can be interpreted as learning constraints justified by the hidden model.
In the present implementation we treat CCA operationally as a conservative
model-refinement procedure rather than as a complete recovery algorithm
for~$T$.

\paragraph{Procedure (Algorithm~\ref{alg:ca}).}
Given a rejected schedule~$e$:
\begin{enumerate}
  \item \textbf{Pair querying.} For each task pair $(i,j)$ whose current
        candidate separations are violated by~$e$, submit a two-task query
        and, if needed, binary-search over the sorted values in $B_{ij}$
        using window-consistent pair assignments. If a strongest justified
        gap $\delta^*$ is found, add $\mathrm{sep}(i,j,\delta^*)$ to~$L$
        and prune all weaker separation candidates for that pair from~$B$.

  \item \textbf{Capacity fallback.} If no justified separation is found,
        learn the weakest violated capacity candidate, namely the smallest
        violated window width together with the largest still-violated
        capacity at that width.
\end{enumerate}
After learning a constraint, prune~$B$ by removing all dominated
candidates. Here $B_{ij}$ denotes the set of candidate separation values
for pair $(i,j)$ currently present in~$B$.

\begin{algorithm}[t]
\caption{CCA: Conservative Constraint Acquisition}
\label{alg:ca}
\begin{algorithmic}[1]
\Require Rejected schedule $e$, learned set $L$, basis $B$,
\For{each pair $(i,j)$ such that $e_i>0$, $e_j>0$, $B_{ij}\neq\emptyset$, and
     $|e_i-e_j| < \max B_{ij}$}
  \State $e' \gets (x_i{=}e_i, x_j{=}e_j, \text{all others}{=}0)$
  \If{$\mathcal{O}(e') = \textit{no}$}
    \State Binary-search the sorted values in $B_{ij}$ for the largest separation
$\delta^*$ whose corresponding pair of slots is rejected by the oracle.
    \If{$\delta^*$ exists}
      \State $L \gets L \cup \{\mathrm{sep}(i,j,\delta^*)\}$
      \State $B \gets B \setminus
             \{\mathrm{sep}(i,j,\delta'') : \delta'' \le \delta^*\}$
      \State \Return \Comment{one constraint per rejection}
    \EndIf
  \EndIf
\EndFor
\State \Comment{No justified separation found $\Rightarrow$ capacity fallback}
\State $w' \gets \min\{w : \exists k,\,
       \mathrm{cap}(k,w)\in B,\; e \not\models \mathrm{cap}(k,w)\}$
\State $k' \gets \max\{k :
       \mathrm{cap}(k,w')\in B,\; e \not\models \mathrm{cap}(k,w')\}$
\State $L \gets L \cup \{\mathrm{cap}(k',w')\}$
\State $B \gets B \setminus
       \{\mathrm{cap}(k'',w'') : k'' \ge k',\; w'' \le w'\}$
\State \Return
\end{algorithmic}
\end{algorithm}
\subsection{Implementation Details}
\label{sec:implementation}

\paragraph{OptSol.}
We implement \textsc{OptSol} using \textbf{CP-SAT} (Google OR-Tools) as a
time-limited anytime solver.  Each call is given a maximum time limit of \texttt{cpsat\_iter\_t} seconds
(20\,s in the reported experiments), although in practice CP-SAT often
terminates much earlier when optimality is certified quickly.  We use
4 parallel workers and a fixed random seed for reproducibility.
CP-SAT returns the best feasible solution found within the time limit,
along with a certificate of optimality if the search completes.  When
optimality is proved, the returned value coincides with the exact optimum
of the current learned model, otherwise it should be
read as the best value found under the current learned model~$L$ within the
time budget.  Accordingly, the experimental ``Opt($L$)'' curves are
empirical CP-SAT values under~$L$, not certified upper bounds unless
CP-SAT has proved optimality at that iteration.

\paragraph{FAO baseline.}
FAO runs CCA for 100~queries, then solves the learned~$L$ with CP-SAT.
If the result violates~$T$, it is repaired by dropping lowest-priority
tasks until oracle-feasible.  Both FAO and L\&O use CCA and receive the
same oracle budget, so the comparison isolates the effect of interleaving
acquisition with optimization.

\subsection{End-to-End Example}
\label{sec:example}

We trace Algorithm~\ref{alg:lo} on a small instance to show how the
interaction works in practice.

\paragraph{Setup.}
Three tasks: $A$ (priority~3), $B$ (priority~2), $C$ (priority~2).
Horizon $H = 10$ slots. Visibility windows $W_A = \{2,3,4\}$,
$W_B = \{3,4,5\}$, $W_C = \{7,8\}$ represent the scheduled-slot part
of each variable's domain: $D(x_j) = \{0\} \cup W_j$.
Hidden constraints (unknown to the algorithm): $\mathrm{sep}(A,B,3)$
(at least 3~slots between $A$ and $B$) and $\mathrm{cap}(1,5)$ (at most
1~task in any 5-slot window).
Candidate basis $B$: $\mathrm{sep}(i,j,\delta)$ with
$\delta \in \{2,3,4\}$ for each pair, plus $\mathrm{cap}(k,w)$ with
$k \in \{1,2\}$, $w \in \{3,4,5\}$.

\paragraph{Initialisation.}
Schedule the highest-priority task: $A$ at slot~3.  The oracle confirms
feasibility.  Set $L = \emptyset$; initialise the basis~$B$ as all
constraints derived from~$\mathcal{L}$ and consistent with $\{A{=}3\}$.
Best known solution: $e^* = \{A{=}3\}$, value $= 3$.

\paragraph{Initial pruning.}
Before the first oracle query, CCA eliminates candidates that can never
be violated.  The minimum slot gap between $A$ and $C$ across their
windows is $\min_{t_A \in W_A,\, t_C \in W_C}|t_A-t_C| = |4-7|=3$, so
$\mathrm{sep}(A,C,2)$ and $\mathrm{sep}(A,C,3)$ are pruned immediately.
Similarly $\mathrm{sep}(B,C,2)$ is pruned ($\min|t_B-t_C|=2\geq 2$).
The basis retains 12 candidates instead of 15.

\paragraph{Iteration 1.}
$\textsc{OptSol}(L{=}\emptyset, f)$ returns
$e = \{A{=}3, B{=}4, C{=}7\}$ (value~7).
Oracle: \textbf{no} ($|3-4|=1 < 3$ violates the hidden separation).

CCA sorts violated pairs by their strongest candidate in $B$.  Pair $(A,B)$ has
$B_{AB} = \{2,3,4\}$ with maximum~$4$ and is queried first.
CCA binary-searches using window-consistent partial queries ($W_A=\{2,3,4\}$,
$W_B=\{3,4,5\}$):
\begin{itemize}
  \item $\delta=3$, gap$=2$: partial query $\{A{=}2, B{=}4\}$, $|2-4|=2$.
        Oracle: \textbf{no} (separation violated: $|2-4|=2<3$).
        Justified; binary search tries $\delta=4$.
  \item $\delta=4$, gap$=3$: partial query $\{A{=}2, B{=}5\}$, $|2-5|=3$.
        Oracle: \textbf{no} (capacity violated: slots~2 and~5 are both
        in window $[1,5]$, count~$=2>1$).  The oracle returns \textit{no}
        regardless of the cause. $\delta=4$ is recorded as justified.
        No stronger candidate remains.
\end{itemize}
The binary search returns $\delta^*=4$.  Because the $\delta{=}4$ partial query
was rejected by the capacity constraint rather than by separation, CCA
learns $\mathrm{sep}(A,B,4)$, an \emph{over-tightened} constraint: the
true hidden separation is $\mathrm{sep}(A,B,3)$, but the algorithm
cannot distinguish the cause of rejection.
Update $L \leftarrow \{\mathrm{sep}(A,B,4)\}$ and
prune all $\mathrm{sep}(A,B,\cdot)$ candidates from~$B$.

\paragraph{Iteration 2.}
With $\mathrm{sep}(A,B,4)$ in~$L$, tasks $A$ and $B$ can never be
co-scheduled: the maximum slot gap achievable in $W_A \times W_B$ is
$|2-5|=3 < 4$.  $\textsc{OptSol}(L, f)$ therefore returns
$e = \{A{=}2, C{=}7\}$ (value~5).
Oracle: \textbf{yes} (no separation constraint on $(A,C)$ and $\mathrm{cap}(1,5)$
is satisfied since slots~2 and~7 lie in disjoint 5-slot windows).
Update $e^* \leftarrow \{A{=}2, C{=}7\}$, value~$=5$.

\paragraph{Outcome.}
Two main oracle queries plus two partial queries.  Best solution:
$\{A{=}2, C{=}7\}$, value~$= 5$.
Learned model: $L = \{\mathrm{sep}(A,B,4)\}$, an over-tightened
constraint of the true $\mathrm{sep}(A,B,3)$. $\mathrm{cap}(1,5)$ is
never queried because the second oracle call returns \textit{yes} before
any capacity violation is encountered.
Fraction of hidden constraints exactly identified: $0/2 = 0$.
The true optimum is~5 and is recovered despite incomplete constraint
learning: the over-tightened separation suffices to steer the solver
toward a feasible optimal schedule, illustrating that exact recovery of
the hidden model is not required for good solutions.

\section{Experiments}
\label{sec:experiments}

\subsection{Setup}

\paragraph{Instance generation.}
We generate instances with $n \in \{10, 20, 30, 40, 50\}$ tasks, planning
horizon $H = 3n$, and priorities drawn from $\mathrm{Uniform}(0.5, 2.0)$.
Each window~$W_j$ is a random sub-interval of $\{1,\ldots,H\}$.  Hidden
constraints: separation on ${\approx}30\%$ of task pairs ($\delta \in
\{2,3,4,5\}$ slots) and one global capacity constraint with window width
$w = \lfloor H/5 \rfloor$ and limit $k$ set empirically so that
Twenty seeds (0--19) per configuration. For
$n \geq 40$ the CP-SAT reference is the best feasible solution found
within 120\,s rather than a proven optimum, so reported gaps are upper
bounds.

\paragraph{Constraint language.}
$\mathcal{L}$ contains $\mathrm{sep}(i,j,\delta)$ for all pairs with
$\delta \in \{2,\ldots,10\}$ (i.e.\ \texttt{sep\_lang\_max}$=10$), plus
$\mathrm{cap}(k,w)$ in a neighbourhood of radius~2 around the true
parameters (${\approx}25$ candidates).  Language sizes range from
${\approx}430$ ($n{=}10$) to ${\approx}11{,}050$ ($n{=}50$).
At initialisation, candidates that are always satisfied given the task
windows (Section~\ref{sec:ca}) are pruned, reducing the effective basis.

\paragraph{Methods.}
\begin{itemize}
  \item \textbf{PG} (Priority Greedy): schedule by descending priority,
        repair by dropping cheapest tasks until oracle-feasible.  Uses no
        constraint knowledge.
  \item \textbf{CP-SAT}: OR-Tools on the full hidden model~$T$ with a
        120\,s limit.  For $n\leq 30$ it proves optimality on all reported
        seeds and for $n\geq 40$ it is treated as a best-feasible reference.
  \item \textbf{FAO} (Full Acquire-then-Optimise): CCA acquisition
        (100~queries) $\to$ CP-SAT on learned~$L$ $\to$ oracle-feasibility
        repair.
  \item \textbf{L\&O}: Algorithm~\ref{alg:lo} with CCA and query cutoff
        $Q = 100$.
\end{itemize}

\paragraph{Metrics.}
\textbf{Gap}: $(f_{\text{ref}} - f) / f_{\text{ref}} \times 100\%$, where
$f_{\text{ref}}$ is the CP-SAT reference value.  
\textbf{Main queries}: number of oracle calls on full schedules (excludes
CCA partial queries).  For FAO this is always 100 and for L\&O it is the
iteration at which the algorithm terminates.
\textbf{$q^*$}: iteration at which the best solution was first found
during the L\&O interleaved phase (not applicable to FAO).
\textbf{Time}: total wall-clock time including the final CP-SAT solve.
\textbf{Speedup}: FAO time / L\&O time.
\textbf{frac}: fraction of~$T$ exactly identified in~$L$ at termination
(reported in the text).

\subsection{Results}

Tables~\ref{tab:instances}--\ref{tab:main} present the results.
L\&O terminates well before the 100-query cutoff on average: the
Accepted-Opt($L$) stopping condition fires after only 5--21 main queries
depending on $n$.  The model-side convergence condition ($f_L^* = f_{L\cup B}^*$)
never fired in any run: with a large candidate basis $B$ and a
time-limited \textsc{OptSol}, the two bounds did not meet within the
query budget before the feasibility of \textsc{OptSol}$(L, f)$ cutoff.

\begin{table}[t]
\centering
\caption{Instance statistics (averaged over 20 seeds).  Separation
  density ${\approx}30\%$ of all task pairs.
  $|B_0|$ is the initial candidate basis size before vacuous pruning and
  $k$ and $w$ are the hidden capacity limit and window width
  of the true constraint $\mathrm{cap}(k,w)$.}
\label{tab:instances}
\small
\begin{tabular}{ccccccc}
\toprule
$n$ & $H$ & $|\mathrm{sep}_T|$ & $k$ & $w$ & $|B_0|$ & CP-SAT ref \\
\midrule
10 & 30  & 13  & 2 & 6  & ${\approx}430$    & optimal \\
20 & 60  & 57  & 2 & 12 & ${\approx}1{,}735$ & optimal \\
30 & 90  & 130 & 4 & 18 & ${\approx}3{,}940$ & optimal \\
40 & 120 & 234 & 5 & 24 & ${\approx}7{,}045$ & best/120s$\dagger$ \\
50 & 150 & 367 & 6 & 30 & ${\approx}11{,}050$ & best/120s$\dagger$ \\
\bottomrule
\end{tabular}
\end{table}

\begin{table}[t]
\centering
\caption{Results averaged over 20 seeds ($\pm$std).
  \textbf{FAO}: 100 acquisition queries + final CP-SAT solve (total time includes the final solve).
  \textbf{L\&O}: cutoff $Q{=}100$, terminating when the current CP-SAT proposal under $L$ is accepted by the oracle.
  Gap: relative to the CP-SAT reference ($\dagger$ = best feasible in 120\,s).
  $q^*$: main query at which the best solution was first found during interleaving.
  Speedup: FAO time / L\&O time.}
\label{tab:main}
\small
\setlength{\tabcolsep}{4pt}
\begin{tabular}{clrrrrr}
\toprule
$n$ & Method & Gap\% & Main queries & $q^*$ & Time & Speedup \\
\midrule
\multirow{2}{*}{10}
 & FAO   & 18.5 (±6.4)  & 100 & ---  & 2.1s   & --- \\
 & L\&O  & \textbf{17.7 (±5.9)} & \textbf{5.3} & 5.3 & \textbf{0.1s}  & \textbf{21$\times$} \\
\midrule
\multirow{2}{*}{20}
 & FAO   & 35.8 (±2.1)  & 100 & ---  & 9.3s   & --- \\
 & L\&O  & 35.8 (±2.1)  & \textbf{8.2} & 8.2 & \textbf{0.8s}  & \textbf{11$\times$} \\
\midrule
\multirow{2}{*}{30}
 & FAO   & 29.5 (±9.7)  & 100 & ---  & 40.7s  & --- \\
 & L\&O  & \textbf{27.1 (±9.3)}  & \textbf{14.2} & 14.2 & \textbf{5.0s}   & \textbf{8$\times$} \\
\midrule
\multirow{2}{*}{40$\dagger$}
 & FAO   & 19.4 (±7.5)  & 100 & ---  & 90.4s  & --- \\
 & L\&O  & 20.2 (±8.1)  & \textbf{16.3} & 16.3 & \textbf{12.8s}  & \textbf{7$\times$} \\
\midrule
\multirow{2}{*}{50$\dagger$}
 & FAO   & 20.3 (±4.8)  & 100 & ---  & 695.0s & --- \\
 & L\&O  & \textbf{17.9 (±7.2)} & \textbf{21.3} & 21.3 & \textbf{130.0s} & \textbf{5$\times$} \\
\bottomrule
\end{tabular}
\end{table}

\subsection{Convergence}
\label{sec:convergence}

\begin{figure}[t]
\centering
\includegraphics[width=0.88\textwidth]{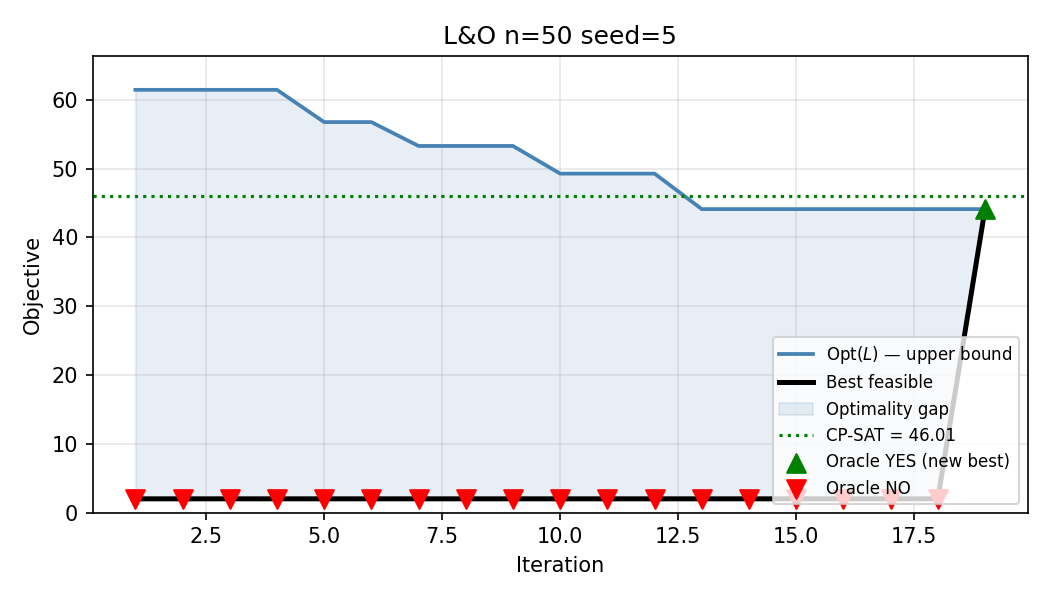}
\caption{Convergence of L\&O on a representative $n{=}50$ instance.
  Red markers indicate oracle rejections, each followed by a CCA update.
  The blue curve is the value returned by the time-limited CP-SAT solve
  under the current learned model~$L$. The black curve is the best
  oracle-confirmed solution found so far. The plot illustrates the typical
  anytime behavior of the current implementation: the constraint-model value
  changes as CCA refines~$L$, while the best accepted solution improves in
  discrete jumps and the run terminates once the current proposal under~$L$
  is accepted by the oracle.}
\label{fig:convergence}
\end{figure}

\paragraph{(1) L\&O substantially improves over no constraint knowledge.}
PG, which ignores the hidden constraints during optimisation, shows average
gaps of 65.0\%, 66.3\%, 68.0\%, 73.4\%, and 72.7\% for
$n\in\{10,20,30,40,50\}$.  L\&O reduces these to 17.7\%, 35.8\%, 27.1\%,
20.2\%, and 17.9\%, respectively.  The improvement is especially clear for
$n\leq 30$, where the CP-SAT reference is proven optimal on all seeds.

\paragraph{(2) L\&O uses far fewer main queries than FAO while remaining competitive in quality.}
The best L\&O solution is found after only 5.3, 8.2, 14.2, 16.3, and
21.3 main queries on average for $n=10,20,30,40,50$, compared with the fixed
100-query acquisition phase of FAO.  In terms of quality, L\&O matches or
improves on FAO at $n=10$, $20$, $30$, and $50$, and is only slightly worse
at $n=40$ (20.2\% vs.\ 19.4\%), a difference well within one standard
deviation ($\pm$8.1 vs.\ $\pm$7.5) and likely attributable to the
stochastic behaviour of time-limited CP-SAT across seeds rather than a
systematic advantage of FAO.  At $n=50$, L\&O improves over FAO on average
by 2.4 percentage points (17.9\% vs.\ 20.3\%), wins on 6 of the 20 seeds,
and reduces average wall-clock time from 695.0\,s to 130.0\,s.

\paragraph{(3) Partial knowledge still suffices.}
The best solution is typically found with only a small fraction of the
hidden constraints exactly identified: the mean value of $\mathrm{frac}$ is
0.096, 0.056, 0.065, 0.045, and 0.041 for $n=10,20,30,40,50$.
This indicates that many hidden constraints never bind the best schedule.
What matters is learning the few constraints that rule out the strongest
competing assignments.

\paragraph{(4) L\&O stops early.}
Figure~\ref{fig:convergence} illustrates a typical pattern.  The CP-SAT value
under the learned model decreases as CCA adds constraints, while the best
oracle-confirmed solution jumps whenever a proposed schedule is accepted.
Because optimization and acquisition are interleaved, L\&O can terminate as
soon as a strong accepted solution is found, rather than spending the full
100-query budget on acquisition before attempting a final solve.  This
anytime property is the dominant practical advantage over FAO in the current
implementation.

\subsection{Discussion}

\paragraph{Why does L\&O help in practice?}
The dominant advantage is not that it perfectly reconstructs the hidden
model, but that it does not wait until the end of acquisition before
searching for good schedules.  FAO always spends its full 100-query budget
before the final CP-SAT call.  In contrast, L\&O typically finds its best
solution after 5--21 main queries and then terminates immediately once an
accepted optimistic proposal is obtained.  This is why the wall-clock
speedups in Table~\ref{tab:main} remain large even when solution quality is
similar.


\paragraph{Partial learning is enough.}
A striking empirical observation is that L\&O does not need to recover most
of the hidden model to find good schedules.  Averaged over seeds,
$\mathrm{frac}$ stays between 0.041 and 0.096.  Note that $\mathrm{frac}$
measures \emph{exact} identification: a constraint learned as
$\mathrm{sep}(i,j,4)$ when the truth is $\mathrm{sep}(i,j,3)$ counts as
unidentified, even though the over-tightened constraint may block the same
competing assignments just as effectively.  What matters in practice is
not exact recovery but learning enough of the constraint structure to
steer the solver away from infeasible high-value schedules.

\paragraph{Oracle call accounting.}
Each L\&O iteration makes one main oracle query, but CCA partial queries add overhead.
The mean total oracle calls grow
from 48 ($n{=}10$) to 727 ($n{=}50$), compared with 143 to 904 for FAO.
If each oracle call corresponds to a simulator run, these totals still
represent minutes rather than the much larger engineering cost of manually
specifying the constraint model.



\section{Conclusion}
\label{sec:conclusion}

We have presented, to our knowledge, the first study of EO satellite
scheduling under unknown operational constraints.  By interleaving
constraint acquisition with optimization, the method learns from oracle
queries while continuously improving a solution schedule.  On a
simplified model with separation and capacity constraints and
dense hidden networks, L\&O improves over a no-knowledge greedy
baseline and uses far fewer main oracle queries than a two-phase
acquire-then-optimise pipeline.  In the current experiments, the strongest
average advantage over FAO appears at $n{=}50$ (17.9\% vs.\ 20.3\% gap,
relative to a best-feasible CP-SAT reference) together with about a
$5\times$ wall-clock speedup.

A key empirical insight is that most hidden constraints never need to be
identified exactly.  The best solution is typically found with only
4--10\% of hidden constraints exactly recovered, suggesting that effective
interactive scheduling depends more on discovering the few constraints that
eliminate strong competitors than on reconstructing the entire hidden
model.

\paragraph{Limitations.}
(i)~CCA may learn over-tightened constraints (as illustrated
in Section~\ref{sec:example}): when a pair query is rejected by a
capacity constraint rather than by separation, CCA records a stronger
separation than the truth. The learned model $L$ may therefore exclude
some feasible schedules, which can cause L\&O to miss the true optimum
even when the CP-SAT solve under $L$ is exact.
(ii)~The experimental implementation uses time-limited CP-SAT inside
\textsc{OptSol}, so the plotted ``Opt($L$)'' values and the stop event
should be interpreted as implementation-level anytime signals rather than
universal optimality certificates.
(iii)~The constraint model is restricted to separation and capacity,
extending to richer families requires either generalizing CCA or replacing
it with a generic algorithm such as \textsc{QuAcq}.
(iv)~We assume a perfect, stationary oracle. Noisy oracles and constraint
drift are not handled.

\paragraph{Future work.}
(i)~Stronger solver engineering for the per-iteration CP-SAT calls,
including better stopping logic
(ii)~validation on real satellite scheduling datasets with physically
grounded constraints (e.g.\ the Eddy \& Kochenderfer
dataset~\cite{aeoseddy2021} with slew-derived separations)
(iii)~multi-satellite constellations
(iv)~noisy and approximate oracles
(v)~constraint drift via periodic basis reset
(vi)~\textsc{QuAcq} for settings where the constraint language is not
known in advance.



\bibliographystyle{plain}

\end{document}